\documentclass{article}
\usepackage{spconf,amsmath,graphicx}
\usepackage{amssymb}
\usepackage{subcaption}

\title{Deep Kinship Verification via Appearance-shape Joint Prediction\\ and Adaptation-based Approach}
\name{Heming Zhang$^1$, Xiaolong Wang$^{2}$, C.-C. Jay Kuo$^1$}
\address{
\begin{tabular}{c c}
$^1$ University of Southern California & $^2$ Samsung Research America\\
Los Angeles, CA, USA & Mountain View, CA, USA\\
\end{tabular}
}%
\begin{document}
%
\maketitle
\begin{abstract}
Kinship verification aims to identify the kin relation between two given face images. It is a very challenging problem due to the lack of training data and facial similarity variations between kinship pairs. In this work, we build a novel appearance and shape based deep learning pipeline. First we adopt the knowledge learned from general face recognition network to learn general facial features. Afterwards, we learn kinship oriented appearance and shape features from kinship pairs and combine them for the final prediction. We have evaluated the model performance on a widely used popular benchmark and demonstrated the superiority over the state-of-the-art.

\end{abstract}

\section{Introduction}
\label{sec:intro}
Human facial image analysis has attracted lots of interests from image processing and computer vision community for a long time, e.g., face recognition~\cite{vgg-face, center-face}, face detection~\cite{finding, scale-aware}, facial attributes perception~(age\cite{age}, gender~\cite{gender}, etc), landmark detection~\cite{landmark}. Compared to general facial image analysis, kinship verification starts to attract our attention recently. There are many potential applications associated with it, such as finding missing children and social data mining~\cite{NRML}. However, identifying the underlying kin relation from facial images is very challenging, even for humans~\cite{prototype}. 

Recently, with the advancement of deep learning technology and big data, we have observed significant improvements in these general facial image analysis problems. Nevertheless, the performance of kinship verification is still not satisfied. One reason is because of the challenge of collecting large kinship data. Unlike general face recognition database collection, we need to collect pairs of facial images with kin relations. Another reason is due to similarity variations among people with kin relations. Even for the same family, e.g., the similarity between the father-son and father-daughter is usually different.  


Most initial work mainly utilizes hand-crafted features, and then learn the prediction with metric learning approach. For example, in~\cite{NRML}, a projection based metric with large margin (NRML) is learned and hand-crafted features such as LBP, HOG, and SIFT are utilized. These features are widely used in many applications, but it is very hard to catch these underlying kin relations. 

Since deep learning has gained success in many fields~\cite{AlexNet, VGG, ResNet}, recent work tend to utilize features from deep neural network (DNN). However, extracting DNN features from limited kinship data is not trivial. It results in either shallow network architectures\cite{smcnn}, or stacked auto-encoder network with hand-crafted features as inputs~\cite{dkv}. In either cases, the power of DNN is degraded which makes it hard to capture these disriminative kin relation features. In~\cite{iccvw2017}, they relabel the kinship data and feed it into a deep neural network to fine-tune the face recognition model, in which large-scale face recognition data is utilized to help the training. Their experiments demonstrated an improvement when combing deep facial features with NRML~\cite{NRML}.



As we know, the data used in kinship verification are also face images~\cite{NRML}. There should be common features between general face recognition and kinship verification. Training deep neural face recognition network via enough face data can utilize the advantage of  deep networks in extracting comprehensive facial appearance features. These features obtained from face recognition task are aimed for person identification. Therefore, they mainly contain salient appearance information that help differentiate people. 

In our case, we not only need the salient appearance information, but also the global appearance and shape information that helps find the underlying kin relation. Therefore, we enforce the feature learning capability via kinship oriented learning framework which has demonstrated the improvement~\cite{smcnn, dkv}. Moreover, we also combine the shape information with the appearance cues together. We are motivated by the observation that facial geometry information is also shared between most kinship face pairs, such as eye shape, nose structure, etc..



\begin{figure*}[htb]
\centering
\includegraphics[trim=130 250 270 460, clip, width=0.7\textwidth]{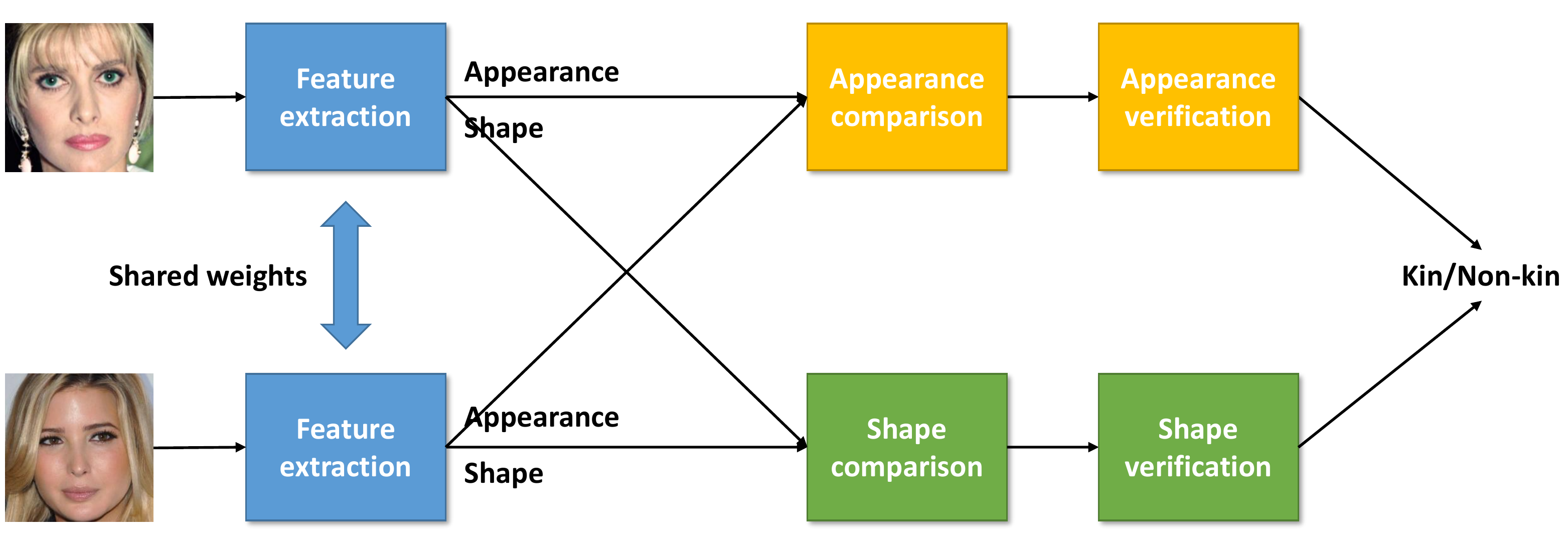}
\caption{Pipeline of our proposed method}
\label{fig:overview}
\end{figure*}


In this work, we utilize both appearance and shape information for kinship verification. Specifically, we design a convolutional neural network for deep appearance and shape feature extraction, comparison and prediction. To overcome the issue with limited kinship data, we propose an adaptation-based two-phase training scheme which utilizes large-scale face recognition data. 

\section{Proposed Method}
\label{sec:method}

\subsection{Overview of proposed method}
In this section, we will talk about the proposed kinship verification scheme in details. The whole framework is demonstrated in Figure~\ref{fig:overview}. Given two facial images, we first extract appearance and shape features for each image separately. Then we extract the facial appearance and shape features for each pair of images by comparing their individual features. 
Afterwards, we fuse appearance and shape results together to predict the kinship relation for the input image pairs.  


The feature extraction module consists of a convolutional neural network as the backbone and the verification modules involve multiple fully connected layers as kinship classifiers. The comparison modules will be explained in details in Section~\ref{sec:comparison_module}. To train this modulized network with limited kinship data, we propose an adaptation-based two-phase training scheme in Section~\ref{sec:training_scheme}.

\subsection{Appearance \& Affine-invariant Shape Comparison}
\label{sec:comparison_module}
In this section, we explain the proposed two feature comparison modules for appearance and shape features. In these modules, we construct features for image pairs using comparison between features individually extracted from single images.\\

\noindent{\bf Appearance Comparison (AC)} \\
The appearance features extracted from the previous module mainly contain essential identification information. To compare appearance features from two different people, we choose to perform element-wise multiplication on two appearance features. It resembles the weighted correlation when combined with fully connected layers in the later stage, and easily fits into current deep learning framework.\\

\noindent{\bf Affine-invariant Shape Comparison (AISC)} \\
The shape of a face can be represented by a set of facial landmarks. However, the geometry of facial landmarks is very sensitive to pose and view variations. Obviously we do not want these changes to affect the comparison performance between two facial shapes. 

Inspired by \cite{grassmann, andy}, we use the Grassmann representation as the affine-invariant shape representation. A Grassmann manifold $G_{m,k}$ is the space in which the points are k planes in $\mathbb{R}^m$ \cite{geometry}. We can associate an $m \times k$ orthogonal matrix $U$ to each $k$-plane $\nu$ in $\mathbb{R}^m$, such that the columns of $U$ form an orthonormal basis for $\nu$.

Given a matrix of coordinates of m facial landmarks $S = [(x_1, y_1); (x2, y2); \cdots ;(x_m, y_m)]$, we can obtain an affine-transformed matrix $S'$ by applying a $2 \times 2$ full rank affine transformation matrix $A$ on the right, i.e. $S' = SA$. It is worth noting that column spaces spanned by $S$ and $S'$ are the same. In other words, the 2-D subspace spanned by $S'=SA$ is invariant to $A$, and thus all $S'$ maps to the same point on the Grassmann manifold. Consequently, the Grassmann representation of $S$ is invariant to affine transformations.

Using the Grassmann representation, the comparison between two shape matrices becomes analyzing the geodesic between two points on the Grassmann manifold. The geodesic between two subspaces represented by projectors $P_0$ and $P_1$ on the Grassmann manifold has the form \cite{geodesic}:

\begin{equation}
P_1 = \exp(tX)P_0\exp(-tX).
\end{equation}

The matrix $X$ can be derived by the eigen-decompsition of $B=P_0-P_1$. Therefore $B$ contains all the information of the geodesic between the two shapes. We can thus use $B$ as the feature for affine-invariant shape comparison as in \cite{grassmann, andy}.\\

\begin{figure*}[htb]
\centering
\includegraphics[trim=130 220 220 120,clip,width=0.7\textwidth]{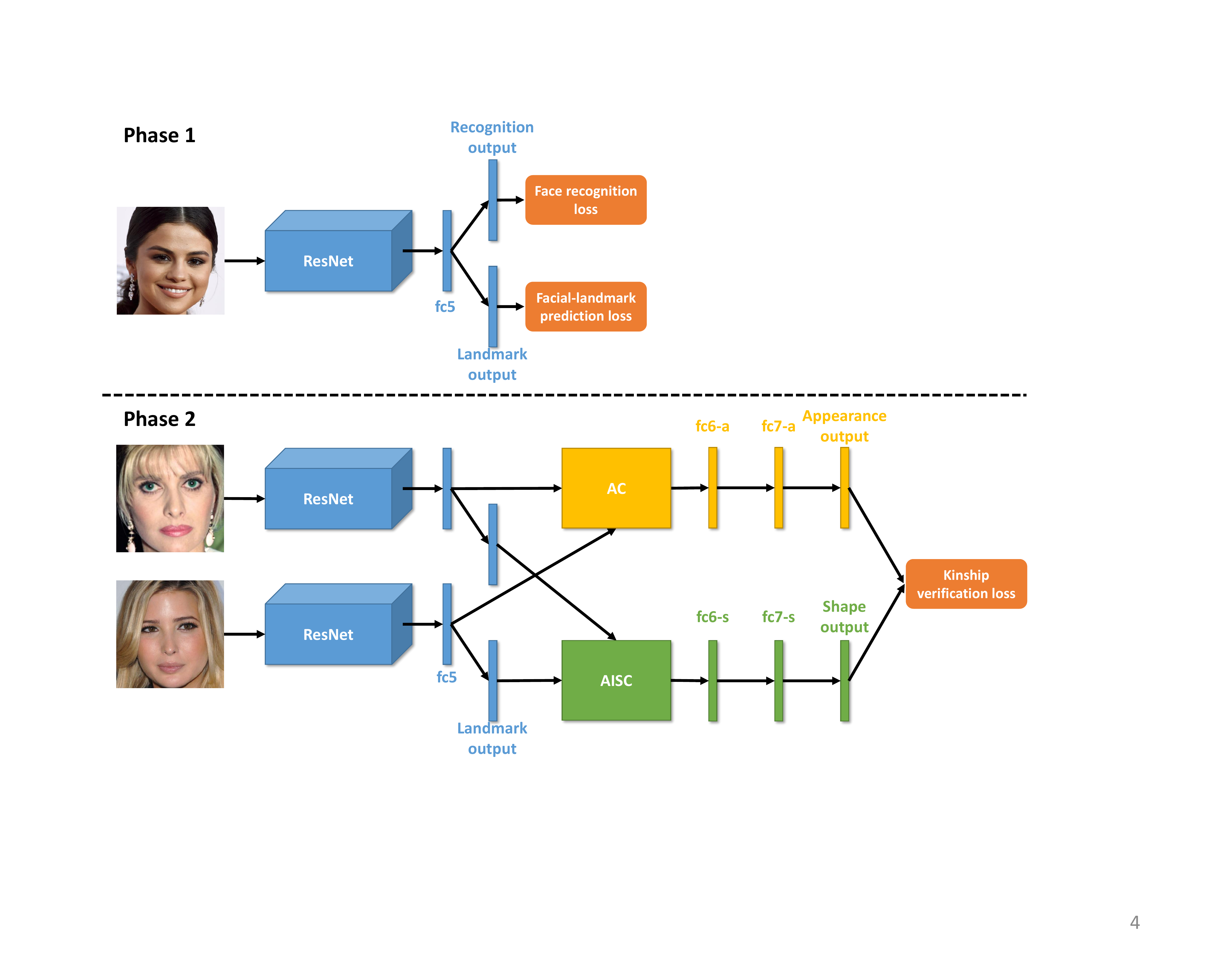}
\caption{Our proposed two-phase training scheme. In phase 1, we utilize large-scale face recognition data to train the feature extraction module. In phase 2, the entire network is retrained on small-scale kinship data.}
\label{fig:two_phases}
\end{figure*}

\noindent{\bf Forward computation of AISC. }
Given two shape matrices $S_0, S_1 \in \mathbb{R}^{m \times 2}$, where each shape matrix contains the $x$ and $y$ coordinates of $m$ landmark points, we compute their SVD as 
\begin{equation}
\begin{split}
S_0 = U_0D_0V_0^T, \;\;
S_1 = U_1D_1V_1^T.
\end{split}
\end{equation}

The corresponding projection matrices are 
\begin{equation}
\begin{split}
P_0 = U_0U_0^T, \;\;
P_1 = U_1U_1^T.
\end{split}
\end{equation}

The shape comparison is conducted as $B = P_0 - P_1$.\\

\noindent{\bf Backward computation of AISC. }
Given the gradient of the loss function $\mathcal{L}$ respective to $B$, the gradient computations for the Grassmann block are given in Eqs. \ref{eq:partial_P} to \ref{eq:partial_S}.
\begin{equation}
\begin{split}
\frac{\partial \mathcal{L}}{\partial P_0} = \frac{\partial \mathcal{L}}{\partial B}, \;\;
\frac{\partial \mathcal{L}}{\partial P_1} = - \frac{\partial \mathcal{L}}{\partial B}. 
\end{split}
\label{eq:partial_P}
\end{equation}

Since the procedures of computing $P_0$ and $P_1$ from $S_0$ and $S_1$ in the forward pass are the same, respectively, we omit the subscript 0 and 1 in the following equations for simplicity.

\begin{equation}
\frac{\partial \mathcal{L}}{\partial U} = 2 \frac{\partial \mathcal{L}}{\partial P} U.
\end{equation}

As proved in \cite{jacobian}, the Jacobian of the SVD can be computed as 
\begin{equation}
\begin{split}
\frac{\partial U}{\partial S_{ij}} = U \Omega_U^{ij}, \;\;
\frac{\partial V}{\partial S_{ij}}  = - V \Omega_V^{ij},
\end{split}
\label{eq:jacobian}
\end{equation}
where $S_{ij}$ is the element on the $i$-th row and $j$-th column of $S$, $\Omega_U^{ij}$ and $\Omega_V^{ij}$ are given by
\begin{equation}
\begin{split}
\Omega_U^{ij} = U^T \frac{\partial U}{\partial S_{ij}}, \;\;
\Omega_V^{ij} = \frac{\partial V}{\partial S_{ij}}^T V.
\end{split}
\end{equation}

The elements of the matrices $\Omega_U^{ij}$ and $\Omega_V^{ij}$ can be computed by solving the linear systems in Eq. \ref{eq:linear_sys}.
\begin{equation}
\begin{cases}
D_l \Omega_{U,kl}^{ij} + D_k \Omega_{V,kl}^{ij} = U_{ik}V_{jl}   \\
D_k \Omega_{U,kl}^{ij} + D_l \Omega_{V,kl}^{ij} = -U_{il}V_{jk}
\end{cases},
\label{eq:linear_sys}
\end{equation}
where $D_l$ is the $l$-th diagonal element in $D$, $\Omega_{U,kl}^{ij}$ is the element on the $k$-th row and $l$-th column of $\Omega_U^{ij}$ and $U_{ik}$ is the element on the $i$-th row and $k$-th column of $U$.

From Eqs. \ref{eq:jacobian} through \ref{eq:linear_sys}, we can derive the gradient for the shape matrix as
\begin{equation}
\begin{split}
\frac{\partial \mathcal{L}}{\partial S_{ij}} &= \sum_m \sum_n \Gamma_{mn},\\
\Gamma &= \frac{\partial \mathcal{L}}{\partial U} \circ (U \Omega_U^{ij}),\\
\Omega_{U,kl}^{ij} &= \frac{D_l U_{ik} V_{jl} + D_k U_{il} V_{jk}}{D_l^2 + D_k^2},
\end{split}
\label{eq:partial_S}
\end{equation}
where $\circ$ denotes the Hadamard product.

\subsection{Adaptation-based Two-phase Training Scheme}
\label{sec:training_scheme}
Most previous work tends to train a shallow network with small-scale kinship data~\cite{smcnn,dkv}. To overcome this issue, inspired by \cite{iccvw2017}, we utilize large-scale face recognition data for training. 

In phase 1, we perform a joint training on face recognition and facial landmark prediction on large-scale face recognition data. The aim of this phase is to train a good feature extraction module for both appearance and shape features. 

In phase 2, we omit the recognition output and add the feature comparison and verification modules to the pre-trained feature extraction module. The network is then trained on small-scale kinship data. The proposed adaptation-based two-phase training scheme for our deep network is illustrated in Figure~\ref{fig:two_phases}.

Comparing to the adaptation approach in \cite{iccvw2017}, our training scheme has two major differences. Firstly, our pre-training in phase 1 involves a joint training on face recognition and facial landmark prediction. This joint training not only helps extract more discriminative appearance features, but also provides extra shape features to further improve the accuracy and robustness of our model. 

Secondly, after the pre-training on large-scale face recognition data in \cite{iccvw2017}, the same network is adopted except for the output layer. To further re-train the face recognition network, each positive kin pair of images are manually assigned with a different identity label. In this way, the network is potentially forced to identify different people as the same person, as well as identify the same person as different people. On the contrary, our modulized network can be adapted to different training data and directly trained on kinship data.

\section{Experiments}
\label{sec:experiments}
\subsection{Experimental Setup}
\noindent{\bf Datasets.} 
To have a fair comparison, we follow the same experimental setting as used in~\cite{iccvw2017} where large-scale face recognition dataset CASIA WebFace is used for adaptation. WebFace contains 500,000 images of 10,000 subjects. For the landmark label, we used dlib~\cite{dlib} to predict the facial landmarks as pseudo labels. We conducted our experiments on the small-scale kinship dataset KinFaceW-I~\cite{NRML}, which contains 1k images. It provides more than 500 pairs of positive and negative samples of four types of kin relationships: father-son (F-S), father-daughter (F-D), mother-son (M-S) and mother-daughter (M-D). \\

\noindent{\bf Training details.} 
The feature extraction module of our network is first trained on WebFace jointly for face recognition and landmark prediction with cross-entropy and mean-squared error as loss functions, respectively. We adopted the ResNet~\cite{ResNet} architecture and trained it from scratch. Then we add the comparison and verification modules and train the network on KinFaceW-I via 5-fold cross validation.

\subsection{Comparison with Previous Work}
Our experimental results on KinFaceW-I are listed in Table~\ref{tab:KinFaceW-I} together with results from previous work. 

One can observe that our proposed method outperforms humans' results~\cite{prototype} as well as handcrafted features~\cite{NRML, prototype}. Comparing with shallow network~\cite{smcnn} or inputs with reduced dimension~\cite{dkv}, we also demonstrate large improvement. 

We achieve the best performance on father-son, father-daughter, mother-son tasks and mean accuracy over four kin relations when comparing with deep network~\cite{iccvw2017}. 

\begin{table}[htb]
\begin{tabular}{|c|c|c|c|c|c|}
\hline
Method & F-S & F-D & M-S & M-D & Mean \\
\hline
Human A \cite{prototype} 	& 62.0 & 60.0 & 68.0 & 72.0 & 65.6 \\
Human B \cite{prototype} 	& 68.0 & 66.5 & 74.0 & 75.0 & 70.9 \\
\hline
MNRML \cite{NRML} 		& 72.5     & 66.5     & 66.2     & 72.0     & 69.6 \\
\hline
MPDFL \cite{prototype}	& 73.5     & 67.5     & 66.1     & 73.1     & 70.1 \\
\hline
DKV \cite{dkv}			& 71.8     & 62.7     & 66.4     & 66.6     & 66.9 \\
\hline
SMCNN \cite{smcnn}		& 75.0     & 75.0     & 68.7     & 72.2     & 72.7 \\
\hline
CFT* \cite{iccvw2017}		& 78.8     & 71.7     & 77.2     & \bf 81.9 & 77.4 \\
\hline
Ours						& \bf 81.8 & \bf 76.6 & \bf 77.5 & 77.2     & \bf 78.3 \\
\hline
\end{tabular}
\caption{Mean verification accuracy (\%) on KinFaceW-I dataset.}
\label{tab:KinFaceW-I}
\end{table}
\begin{table}[htb]
\begin{tabular}{c|ccc|c}
\hline
Setting & Joint pre-training & Appearance & Shape & Mean \\
\hline
A &         & $\surd$ &         &  65.9\\
B & $\surd$ & $\surd$ &         &  69.6\\
C & $\surd$ &         & $\surd$ &  68.0\\
D & $\surd$ & $\surd$ & $\surd$ & \bf 78.3 \\
\hline
\end{tabular}
\caption{Ablation study}
\label{tab:ablation}
\end{table}

\subsection{Ablation Study}
To investigate the contribution to the performance of each component, we conducted an ablation study. We provide the results in Table~\ref{tab:ablation}, in which the setting D is our proposed setting. 

The comparison between setting A and B demonstrates the benefit of joint pre-training with landmark objective, which also enables the utilization of shape information. By comparing settings B, C, D, one can find out that the appearance and shape information are complementary to each other and thus the fusion achieves a significant improvement in performance. 
Without the utilization of both appearance and shape information, our adaptation-based approach performs worse than the previous adaptation-based approach~\cite{iccvw2017}. The reason is that we used a much deeper backbone network structure compared to that in~\cite{iccvw2017}. Consequently, without training both the appearance and shape branches, the network is easily overfitting to the limited kinship data.

\section{Conclusion}
\label{sec:conclusion}
Kinship verification is a challenging task on which even human cannot perform well. In this work we proposed to use deep learning techniques for its promising performance. We utilized two types of complementary information, namely appearance and shape. Both of them are essential for identifying the underlying kin-relation from facial images. To enable the training of such deep network with limited kinship data, we proposed an adaptation-based two-phase training scheme and utilized large-scale face recognition data. We demonstrated the superiority of our proposed method over the state-of-the-arts. 

%


\bibliographystyle{IEEEbib}
\bibliography{ref}
\end{document}